\documentclass[11pt]{article}

\usepackage[preprint]{acl}

\usepackage{times}
\usepackage{latexsym}
\usepackage{booktabs}
\usepackage{multirow}
\usepackage{amsmath}
\usepackage{algorithm}
\usepackage{algpseudocode}
\usepackage[most]{tcolorbox}
\usepackage{float}
\usepackage{amsmath}
\usepackage{amssymb}
\usepackage{bm}
\usepackage[T1]{fontenc}

\usepackage[utf8]{inputenc}

\usepackage{microtype}

\usepackage{inconsolata}

\usepackage{graphicx}

\title{EASE-TTT: Evidence-Aligned Selective Test-Time Training for Long-Context Question Answering}

\author{Xiaopeng Yuan$^{1}$, Zebin Wang$^{2}$, Suwen Wang$^{3}$, Zongxin Yang$^{2}$, Haohan Wang$^{1}$, Yushun Dong\thanks{Corresponding author: yushun.dong@fsu.edu}$^{4}$ \\ $^{1}$University of Illinois Urbana-Champaign \\ $^{2}$Harvard University \\ $^{3}$Brion, ASML US LP \\ $^{4}$Florida State University \\ }

\begin{document}
\maketitle
\begin{abstract}
Long-context question answering (QA) remains challenging for smaller language models even when answer-bearing evidence is already present in the input. Existing within-context retrieval methods localize and expose candidate evidence chunks for the question, but they stop at input-level evidence exposure rather than adapting the query-side attention parameters that control how the model allocates attention over full-context positions. In contrast, lightweight test-time adaptation methods, such as query-only test-time training (qTTT), leave evidence localization unresolved because their generic span-level self-supervised objectives do not identify which context positions support the current answer. In this paper, we propose \underline{E}vidence-\underline{A}ligned \underline{SE}lective \underline{T}est-\underline{T}ime \underline{T}raining (\textbf{EASE-TTT}), a within-context retrieval-augmented test-time training framework that converts selected evidence chunks into a soft attention supervision target over their token positions. Instead of replacing the full context with retrieved chunks, EASE-TTT uses the resulting attention target to guide query-side adaptation, with the adapted model generating the final answer from the original full context. Experiments on six LongBench QA tasks and three small decoder-only language models show that EASE-TTT achieves the strongest macro-average performance among full-context inference, retrieval-only baselines, and qTTT, supporting evidence-aligned test-time adaptation in long-context QA.
\end{abstract}

\section{Introduction}

Large language models have made rapid progress in extending their context windows, enabling them to process inputs that contain tens or even hundreds of thousands of tokens\citep{ding2024longrope, team2024gemini, chen2024longlora}. However, a longer context window does not necessarily translate into better long-context question-answering performance. In many long-context question answering tasks, the answer-bearing evidence is already present in the input, yet the model still fails to access it correctly\citep{liu2024lost, hsieh2024ruler, modarressi2025nolima}. This issue is particularly important for smaller language models, which often have more limited capacity to maintain reliable evidence use in long, distractor-heavy contexts~\citep{gao2026u}. In such cases, the bottleneck is not simply whether the model can fit the context, but whether it can reliably access and prioritize the evidence needed for the current question.

\begin{figure}[t]
\centering
\includegraphics[width=\columnwidth]{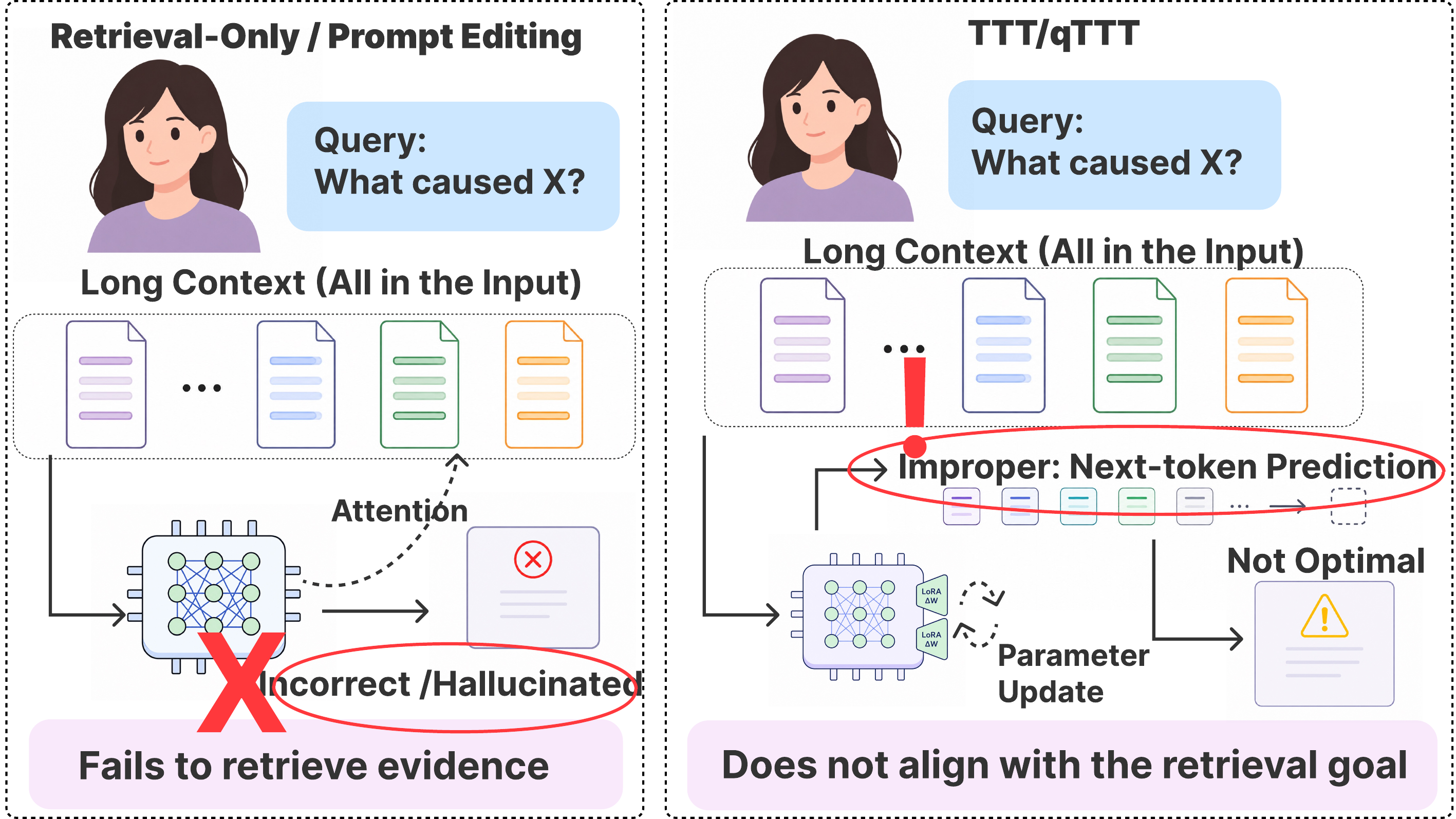}
\caption{Motivation of EASE-TTT. Retrieval-only and prompt-editing methods expose candidate evidence at the input level, but do not adapt the model's context-access behavior. Test-time training methods can adapt model parameters at inference time, but their objectives are often not explicitly aligned with question-relevant evidence. EASE-TTT bridges this gap by using retrieved evidence to guide test-time adaptation.}
\label{fig:motivation}
\end{figure}

A natural way to address this issue is to perform retrieval within the input context. Within-context retrieval methods segment the long input into chunks, localize candidate evidence chunks from the same context, and use the selected chunks to construct a shorter or more focused input~\citep{jiang2024longllmlingua, li2023compressing, nair2023drilling}.  These methods do not rely on an external corpus; instead, they treat the given long context itself as the retrieval source. They are effective when the selected chunks contain sufficient answer-bearing evidence for generation. However, they typically use retrieval only as an input-level operation: selected chunks are used to replace, shorten, or prepend to the original context~\citep{sheng2025dynamic, liskavets2025prompt, wang2023learning, chirkova2025provence}. As a result, the model's parameters and context-access behavior remain unchanged. Moreover, hard chunk selection may discard useful surrounding information, which is risky in long-context QA where evidence may be distributed across multiple parts of the input~\citep{sarthi2024raptor, tian2025distance, saad2024benchmarking, luo2025does, wang2024leave}.

\begin{figure*}[t]
    \centering
    \includegraphics[width=\textwidth]{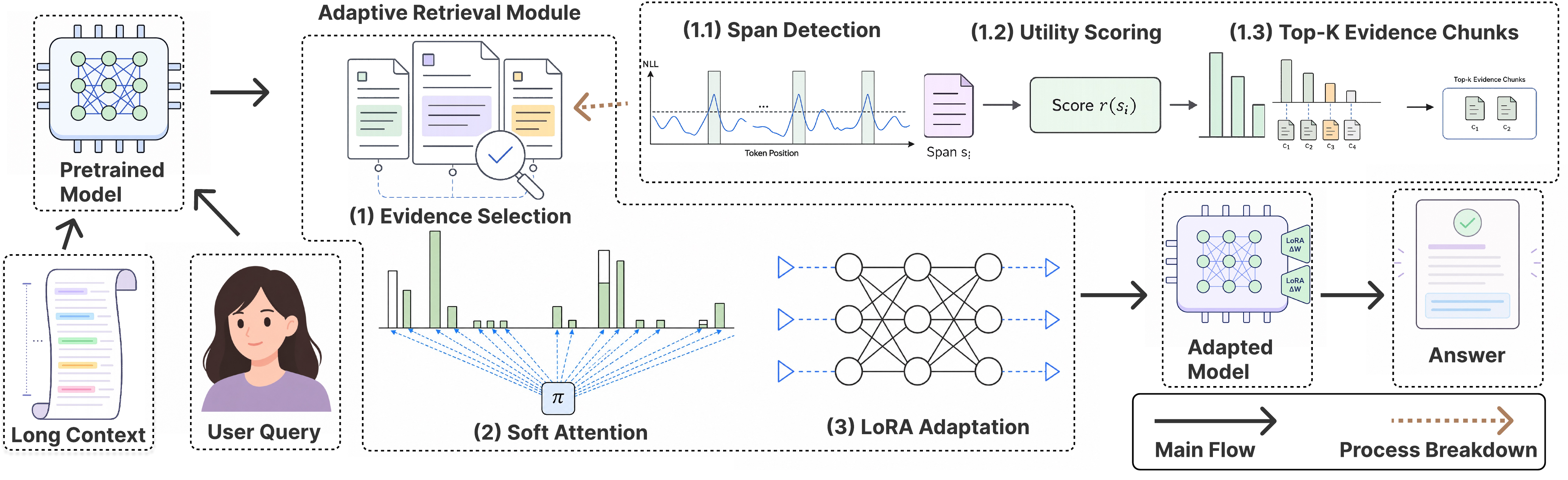}
    \caption{Overview of EASE-TTT. Given a long context and a question, EASE-TTT selects question-relevant evidence chunks, converts them into a soft attention target over full-context positions, and updates query-side LoRA adapters at test time. The adapted model then generates the answer from the original full context.}
    \label{fig:workflow}
\end{figure*}

This limitation suggests that evidence access should not be treated only as an inference-time input selection problem. For smaller models in particular, failures under long contexts may reflect a mismatch between the model's current context-access behavior and the evidence required by the question~\citep{zhu2025focus, lee2025ethic, an2024make, li2024long}. Test-time adaptation provides a natural way to address this mismatch because it allows a model to change its behavior for each test instance at inference time. In this work, we focus on test-time training (TTT), a gradient-based form of test-time adaptation that performs instance-specific parameter updates~\citep{sun2020test, wang2020tent, hardt2024test, akyurek2024surprising}. Recent query-only test-time training further shows that inference-time compute need not be spent only on additional generated tokens; it can also be used for query-side adaptation, allowing the model to change how it allocates attention over a given long context~\citep{bansal2025let}. This perspective is especially relevant to long-context QA, where the evidence may already be present in the input but insufficiently prioritized by the model. However, existing test-time adaptation objectives are typically driven by generic self-supervised, task-level, or retrieval-oriented signals, rather than evidence-localized supervision that identifies which full-context positions support the current answer~\citep{zhang2024come, feng2026place, jeong2023test, sun2026predict}. These objectives may adapt the model to the current input, but they do not explicitly indicate which context positions support the current answer. Therefore, there remains a gap between within-context evidence localization and test-time adaptation: within-context retrieval can localize potentially relevant chunks, while query-side test-time training can adapt model behavior, but existing methods do not directly use question-relevant evidence as supervision for instance-specific adaptation.

We propose \textbf{Evidence-Aligned Selective Test-Time Training (EASE-TTT)}, a within-context retrieval-augmented test-time training framework that turns question-relevant evidence into direct supervision for long-context adaptation. Given a long-context question answering instance, EASE-TTT first selects chunks in the input context that are most relevant to the question. Instead of replacing the original context with these chunks, it constructs a soft attention target that assigns greater probability mass to selected evidence positions while still preserving nonzero mass over the remaining context. At test time, EASE-TTT updates lightweight query-side adapters with the base model frozen. After adaptation, the model generates the answer from the original full context. This design turns retrieval from an input-filtering mechanism into an evidence-aligned supervision signal for instance-specific adaptation.

\paragraph{Our contributions.}
\begin{itemize}
    \item We identify evidence-use failure as a key bottleneck in long-context reasoning for smaller language models: relevant evidence may be present in the input, but the model still fails to use it under distractor-heavy contexts.
    
    \item We propose EASE-TTT, a within-context retrieval-augmented test-time training framework that converts question-relevant chunks into soft supervision for query-side adaptation. Unlike retrieval-only methods, EASE-TTT does not replace the context with chunks; instead, it uses them to guide adaptation while preserving full-context generation.
    
    \item We conduct an evaluation on long-context QA benchmarks across multiple small language models. Our results show that EASE-TTT improves answer quality over full-context inference, retrieval-only baselines, and qTTT, with further analyses demonstrating the effects of evidence selection, soft attention supervision, and test-time training.
\end{itemize}

\section{Related Work}
\textbf{Within-Context Retrieval and Evidence Selection.}
A common approach to long-context question answering is to localize question-relevant evidence within the input context before generation~\citep{li2024alr, qiu2025eliciting, lee2024human}. Unlike standard retrieval-augmented generation~\citep{lewis2020retrieval}, which retrieves passages from an external corpus, within-context retrieval treats the given long input itself as the retrieval source~\citep{qian2024grounding, taguchi2025efficient}. Prior work has explored related strategies such as prompt compression, context pruning, discourse-based document selection, and hierarchical retrieval to reduce distractors and expose useful evidence to the model~\citep{jiang2023llmlingua, zhao2024longrag, yoon2024compact}. Efficiency-oriented variants also rely on selecting, compressing, or reorganizing input passages before generation~\citep{xu2023recomp, pan2024llmlingua}. However, these methods treat evidence access mainly as an input-level operation: retrieved chunks are used to replace, shorten, reorder, or prepend to the original context. As a result, the model's parameters and context-access behavior remain unchanged. This is limiting when answer-bearing evidence is already present in the context window but is still not reliably accessed by the model. Moreover, hard selection can introduce a new bottleneck: selected chunks may omit useful surrounding context, separate evidence distributed across distant regions, or remove information needed to interpret the retrieved span~\citep{gunther2024late, tian2025distance}. Thus, retrieval and prompt editing can change what the model sees, but they do not change how the model attends to and uses evidence in the full context.

\noindent \textbf{Test-Time Training.}
Test-time training (TTT) improves model behavior at inference time by updating parameters using self-supervised signals derived from the test input itself~\citep{hu2025test, zhang2025survey}. These approaches have been explored in settings such as distribution shift, domain adaptation, and reasoning-time adaptation, where fixed pretrained parameters may be insufficient for the input at hand~\citep{hubotter2025efficiently, agarwal2025first, li2025learning}. In the long-context setting, TTT is especially relevant because each test instance may exhibit different local structures, evidence layouts, and distraction patterns~\citep{muhtar2024streamadapter}. However, parameter-level adaptation alone does not solve evidence access unless the training signal is aligned with the evidence required by the current question. Applying TTT to long contexts is therefore nontrivial: the adaptation signal is often local, partial, and potentially noisy, while broad parameter updates may introduce instability or unnecessary computational overhead~\citep{su2023beware, zhang2024come}. These challenges make targeted and evidence-aligned test-time training important for long-context inference. Query-only test-time training (qTTT) narrows the update to the query projections in self-attention rather than adapting the full model~\citep{bansal2025let}. However, qTTT still relies on generic self-supervised objectives rather than explicit supervision from question-relevant evidence. As a result, it can update query-side attention parameters, but it does not specify which full-context positions should guide the update. This creates a mismatch for long-context QA: the model is adapted, but the adaptation is not anchored to the evidence needed to answer the question.

\noindent \textbf{Gap and Motivation.}
These two lines of work address different sides of the long-context evidence-access problem, but neither resolves it alone. Within-context retrieval and prompt editing operate at the input level: they can localize or expose candidate evidence, but they leave the model's context-access behavior unchanged. This is insufficient when the relevant content is already inside the context window but the model fails to attend to it. Query-only test-time training operates at the parameter level: it can adapt query-side attention behavior, but its objectives are not tied to the evidence positions required by the current question. Consequently, existing methods either select evidence without adapting the model, or adapt the model without explicit evidence guidance. Our method bridges this gap by using retrieved evidence chunks not as a replacement for the full context, but as supervision for query-side test-time training. The final answer is still generated from the original full context, while the retrieved evidence guides how the model updates its attention behavior.

\section{Preliminary}
\subsection{Long-Context Question Answering and Evidence Use}

We study test-time training for long-context question answering. Let a test instance be $z=(c,q)$, where $c=(c_1,c_2,\dots,c_T)$ denotes a long input context and $q$ denotes the question or instruction. Given a pretrained language model $f_\theta$, the goal is to generate an answer $y$ conditioned on both the full context $c$ and the question $q$, i.e., $y \sim p_\theta(\cdot \mid c,q)$.

In long-context QA, the relevant evidence needed to answer $q$ may already be contained in $c$, but the model may still fail to identify or use it correctly. This failure is especially problematic when the context contains many distractors or when the useful evidence is distributed across distant regions of the input. Therefore, the key challenge is not only whether the model can fit the full context, but whether it can reliably access the evidence needed for the current question.

A common way to improve evidence access is to perform retrieval within the given context. Let $\mathcal{S}=\{s_1,s_2,\dots,s_M\}$ denote a set of candidate chunks segmented from $c$, where each chunk $s_j=(c_{b_j},\dots,c_{e_j})$ covers a contiguous span of context tokens. A within-context retrieval module ranks these chunks according to their relevance to $q$ and selects a subset $E=\{s_{j_1},s_{j_2},\dots,s_{j_K}\}$. Retrieval-only methods typically use $E$ to construct a shorter input for generation. In contrast, our goal is not to replace the original context with the selected chunks. Instead, we use the selected evidence chunks as a supervision signal for test-time adaptation, while final answer generation remains conditioned on the original full context $c$.

\subsection{Query-Only Test-Time Adaptation}

Test-time training adapts a model independently for each test instance at inference time, using signals derived from the test input itself. In the long-context setting, full-parameter adaptation is expensive because each gradient update may change the key and value representations of the entire context, requiring repeated computation over the full input.

Query-only test-time training provides a lightweight alternative. Instead of updating all model parameters, it updates only query-side parameters in self-attention while keeping the rest of the model frozen. Let \(\Theta_Q = \{W_Q^{(1)}, W_Q^{(2)}, \dots, W_Q^{(L)}\}\) denote the query projection parameters across the \(L\) transformer layers. Given the long context \(c\), the model constructs key-value representations \(\{K^{(\ell)}, V^{(\ell)}\}_{\ell=1}^{L}\), which remain fixed during adaptation. Updating only \(\Theta_Q\) changes how the model forms queries over these fixed key-value representations, thereby modifying how it accesses information in the context without recomputing the full context after every gradient step.
Standard query-only test-time training usually relies on generic self-supervised objectives. For example, it may sample a span $s=(c_t,c_{t+1},\dots,c_{t+m})$ from the context and optimize a next-token prediction loss:
\[
\mathcal{L}_{\mathrm{span}}(\Theta_Q; s)
=
-\sum_{i=t}^{t+m-1}
\log p_{\theta,\Theta_Q}(c_{i+1} \mid c_{\leq i}).
\]
This objective can adapt the model to the current input, but it does not explicitly indicate which parts of the context are useful for answering the current question. As a result, query-only adaptation can modify context-access behavior, but the adaptation signal remains largely question-agnostic.

\subsection{From Evidence Selection to Adaptation Supervision}

The above discussion suggests a gap between within-context retrieval and query-only test-time adaptation. Within-context retrieval can identify candidate evidence chunks for the current question, but retrieval-only methods usually use these chunks to modify the input rather than the model. Query-only test-time training can adapt how the model accesses the context, but its generic span-based objectives do not directly specify which context positions are question-relevant.

Our method connects these two components by using retrieved evidence chunks as supervision for query-side test-time adaptation. Let \(E\) denote the selected evidence chunks, and let \(\Omega(E) \subseteq \{1,2,\dots,T\}\) denote the indices of context tokens covered by these chunks. Instead of replacing the original context with \(E\), we use \(\Omega(E)\) to guide adaptation toward evidence-bearing positions. The detailed construction of the soft attention target and the corresponding adaptation objective are introduced in the next section.

\section{EASE-TTT}
\label{sec:method}

\subsection{Method Overview}
We propose \textbf{EASE-TTT}, an evidence-selective variant of query-only test-time training for long-context question answering. Unlike prior qTTT methods, which adapt query-side parameters using generic self-supervised losses over randomly sampled spans, our method identifies question-relevant evidence and uses it to guide test-time attention adaptation. Given a context \(c\) and a question \(q\), EASE-TTT segments the context into candidate spans and ranks them by their question-conditioned utility. The top-\(K\) spans are selected as evidence chunks and used to define a soft target attention distribution over context positions. During test-time adaptation, EASE-TTT updates only query-side adaptation parameters according to this evidence-aligned attention target. Final prediction is still performed on the original full context, so the selected chunks guide attention without truncating the input.

\begin{algorithm}[t]
\caption{EASE-TTT with Evidence Selection and Soft Attention Supervision}
\label{alg:selective_qttt}
\begin{algorithmic}[1]
\Require Base model $f_\theta$, context $c$, question $q$, update steps $N$, top-$K$, attention layer $\ell$, mass $\alpha$, learning rate $\eta$
\State Insert trainable LoRA adapters into query projections; freeze all other parameters
\State Segment the context into candidate spans $\mathcal{S}$
\For{each span $s \in \mathcal{S}$}
    \State Compute question-conditioned utility score $r(s)$
\EndFor
\State $E \gets \mathrm{TopK}(\mathcal{S}, r, K)$ \Comment{selected evidence chunks}
\State $\Omega(E) \gets \{\text{context token positions covered by } E\}$
\State Construct soft target distribution $\pi$ over context positions using $\Omega(E)$ and $\alpha$
\For{$t = 1$ to $N$}
    \State Obtain attention distribution $a$ over context positions at layer $\ell$
    \State $\mathcal{L} \gets D_{\mathrm{KL}}(\pi \,\|\, a)$
    \State Update query-side LoRA parameters with learning rate $\eta$
\EndFor
\State Generate the final answer using the full context
\end{algorithmic}
\end{algorithm}

\subsection{Within-Context Evidence Selection}
A central challenge in long-context reasoning is that useful evidence is often buried among large amounts of irrelevant content. To obtain a more targeted adaptation signal, we first identify candidate evidence chunks from the full context.

Given the context token sequence \(c=(c_1,\dots,c_T)\), we segment it into spans using token-level negative log-likelihood (NLL) spikes. Specifically, we run a forward pass over \(c\) and compute the NLL of each context token. After smoothing the resulting NLL curve, we detect boundary candidates using a threshold of the form \(\mu+\kappa\sigma\), where \(\mu\) and \(\sigma\) are the mean and standard deviation of the smoothed curve, and \(\kappa\) is a spike factor. Together with a minimum chunk-length constraint \(m_{\min}\), this yields a set of candidate spans \(\mathcal{S}=\{s_1,\dots,s_M\}\).

We then score each span by how much it helps the model condition on the question. For a candidate span \(s\), we define its question-conditioned utility as
\[
r(s)
=
\mathcal{L}_{\text{NTP}}([{\rm BOS}, q])
-
\mathcal{L}_{\text{NTP}}([s, {\rm BOS}, q]),
\]
where $\mathcal{L}_{\text{NTP}}(\cdot)$ denotes the next-token prediction loss on the question tokens. Intuitively, if prepending $s$ reduces the question modeling loss, then $s$ likely contains evidence relevant to answering $q$.

We rank all spans by $r(s)$ and retain the top-$K$ spans:
\[
E=\operatorname{TopK}(\mathcal{S}, r, K),
\]
where \(E\) denotes the selected evidence chunks. These chunks are not used to replace the full context at inference time; instead, they provide a focused supervision signal for the subsequent adaptation stage.

\subsection{Soft-Target Attention Alignment}
Existing qTTT methods typically optimize generic self-supervised objectives such as next-token prediction over sampled spans. While lightweight, such objectives only indirectly encourage the model to allocate attention toward question-relevant evidence. To make the adaptation target more explicit, we supervise attention directly using the selected evidence chunks.

Let \(q=(q_1,\dots,q_R)\) denote the tokenized question. At each test-time adaptation step, we prefill the model on the sequence \([c; q_{1:R-1}]\) and decode the final question token \(q_R\). From a chosen attention layer \(\ell\), we extract the attention distribution over context positions, average across heads, and normalize it into a probability distribution \(a \in \mathbb{R}^{T}\).

Let \(\Omega(E)\) be the set of context token positions covered by the selected evidence chunks \(E\). We define a soft target attention distribution $\pi$ over context positions by assigning most of the probability mass to \(\Omega(E)\):
\[
\pi_i=
\begin{cases}
\alpha/|\Omega(E)|, & i\in\Omega(E),\\[4pt]
(1-\alpha)/(T-|\Omega(E)|), & i\notin\Omega(E),
\end{cases}
\]
where $\alpha\in(0,1)$ controls how strongly attention is biased toward the selected evidence.

We then optimize the Kullback--Leibler divergence
\[
\mathcal{L}_{\text{attn}}
=
D_{\mathrm{KL}}(\pi \| a),
\]
which explicitly encourages the model to reallocate attention toward evidence-bearing context positions while still preserving a small amount of mass on the rest of the context. Compared with hard masking, this soft target is more stable and avoids forcing the model to ignore potentially useful non-selected tokens entirely.

\section{Experiments}
\begin{table*}[ht]
\centering
\small
\caption{Main results on six LongBench QA tasks: MuSiQue, HotpotQA, 2WikiMultihopQA, QASPER, NarrativeQA, and MultiFieldQA-en, across Qwen3-0.6B, Qwen3-1.7B, and Llama-3.2-1B. RAG denotes Within-Context Retrieval-Augmented Generation.}
\label{tab:main_results}
\renewcommand{\arraystretch}{1.2}
\setlength{\tabcolsep}{5pt}
\begin{tabular}{llccccccc}
\toprule
Model & Method & MuSiQue & HotpotQA & 2WikiMQA & QASPER & NarrativeQA & MultiFieldQA & Avg. \\
\midrule

\multirow{5}{*}{Qwen3-0.6B}
& Full-context & 8.0 & 17.9 & 17.0 & 26.0 & 9.6 & 38.8 & 19.5 \\
& RAG          & 7.2 & 17.9 & 17.2 & 26.2 & 11.6 & 37.6 & 19.6 \\
& ICR         & 6.3 & 21.3 & 21.4 & 11.3 & 9.8 & 38.5 & 18.1 \\
& qTTT         & 8.9 & \textbf{23.6} & 20.2 & 29.8 & 12.1 & 39.5 & 22.4 \\
& Ours         & \textbf{9.2} & 23.5 & \textbf{22.1} & \textbf{32.1} & \textbf{14.0} & \textbf{40.4} & \textbf{23.6} \\
\midrule

\multirow{5}{*}{Qwen3-1.7B}
& Full-context & 12.4 & 30.4 & 22.1 & 29.9 & 16.1 & 39.1 & 25.0 \\
& RAG          & 12.7 & 31.4 & 22.7 & 28.9 & \textbf{16.7} & 39.1 & 25.3 \\
& ICR          & \textbf{16.0} & 33.6 & 31.7 & 27.1 & 13.8 & 43.3 & 27.6 \\
& qTTT         & 13.6 & 33.4 & 28.2 & 37.2 & 16.2 & 43.3 & 28.7 \\
& Ours         & 14.9 & \textbf{36.6} & \textbf{32.1} & \textbf{39.2} & 16.3 & \textbf{44.6} & \textbf{30.6} \\
\midrule

\multirow{5}{*}{Llama-3.2-1B}
& Full-context & 11.1 & 19.1 & \textbf{28.9} & 16.2 & 14.0 & 28.9 & 19.7 \\
& RAG          & 9.9 & 22.0 & 28.2 & 21.0 & 16.5 & 33.9 & 21.9 \\
& ICR          & 15.1 & 25.3 & 26.6 & 15.7 & 17.1 & 39.8 & 23.3 \\
& qTTT         & \textbf{15.4} & 27.1 & 26.3 & \textbf{24.8} & \textbf{17.3} & 40.8 & 25.3 \\
& Ours         & 13.2 & \textbf{29.3} & 26.5 & 24.3 & 16.0 & \textbf{45.6} & \textbf{25.8} \\
\bottomrule
\end{tabular}
\end{table*}
\subsection{Setup}

\textbf{Evaluation Datasets.}
We evaluate our method on six English long-context question answering tasks from LongBench~\citep{bai2024longbench}: MuSiQue, HotpotQA, 2WikiMultihopQA, QASPER, NarrativeQA, and MultiFieldQA-en. These tasks cover multi-hop question answering, single-document question answering, narrative understanding, and long-context information extraction. They require models to locate, aggregate, and reason over relevant evidence in extended input contexts. We report the official task-level evaluation scores and compute the macro-average across the six datasets.

\noindent \textbf{LLMs and Baselines.}
We conduct experiments on three small decoder-only language models: Qwen3-0.6B, Qwen3-1.7B ~\citep{yang2025qwen3}, and Llama-3.2-1B ~\citep{grattafiori2024llama}. We compare EASE-TTT with four baselines.
\textbf{Full-context} directly generates the answer from the full input context, without retrieval or test-time parameter updates.
\textbf{Within-Context Retrieval-Augmented Generation (Within-Context RAG)} retrieves the top-ranked chunks from the same input context using the question as the retrieval query, concatenates the retrieved chunks as a shortened context, and generates the answer without accessing any external corpus or updating model parameters.
\textbf{In-Context Retrieval (ICR)} retrieves relevant segments from the given long input and uses the retrieved segments, together with the corresponding prompting strategy, to answer the question~\citep{agrawal2024can}.
\textbf{Query-Only Test-Time Training (qTTT)} performs query-only test-time training by updating query-side parameters using a generic self-supervised next-token prediction objective on sampled context spans~\citep{bansal2025let}.
\textbf{EASE-TTT} updates only query-side adaptation parameters, but replaces generic span-based supervision with evidence-guided soft attention supervision constructed from question-relevant chunks selected within the input context. Unlike retrieval-only baselines, EASE-TTT uses selected chunks only to guide test-time adaptation, while final answer generation is performed over the original full context.

\noindent \textbf{Implementation Details.}
For all methods, we truncate the input context to at most 32,768 tokens and the question to at most 1,024 tokens. The maximum answer length is set to 128 tokens, and we use deterministic decoding. For EASE-TTT, we insert LoRA adapters into the query projection modules while keeping the base model frozen. Unless otherwise specified, we use LoRA rank 8, a scaling factor of 16, and a dropout rate of 0.05. Test-time adaptation is performed for 15 update steps with AdamW, using a learning rate of $1\times10^{-4}$ and weight decay of 0.01. We use 512 tokens as the target chunk size, with a minimum chunk size of 128 tokens, a maximum chunk size of 1,024 tokens, and an overlap of 64 tokens. We then rank candidate chunks by the utility score in Section~\ref{sec:method} and select the top 4 chunks for evidence-guided adaptation. By default, we use layer $\ell=14$ for attention alignment. The soft attention target uses a mass parameter of $\alpha=0.6$.

\subsection{Main Results}
Table~\ref{tab:main_results} reports the main results on six LongBench QA tasks. Overall, EASE-TTT achieves the best average performance on the Qwen3 models, improving over full-context inference, retrieval-only baselines, and qTTT. On Qwen3-0.6B, EASE-TTT obtains an average score of 23.6, outperforming full-context inference by 4.1 points and qTTT by 1.2 points. On Qwen3-1.7B, EASE-TTT achieves an average score of 30.6, improving over Full-context by 5.6 points, Within-Context RAG by 5.3 points, ICR by 3.0 points, and qTTT by 1.9 points.

These results support our hypothesis that long-context QA depends not only on context availability, but also on reliable evidence access. Full-context inference is consistently weaker than adaptation-based methods, indicating that simply providing the full input is insufficient. Retrieval-only methods improve some tasks, but their gains are inconsistent. For example, ICR improves MuSiQue and 2WikiMultihopQA on Qwen3-1.7B, but underperforms full-context inference on QASPER and NarrativeQA, suggesting that shortened retrieved contexts may also discard useful surrounding information.

Compared with qTTT, EASE-TTT improves the macro-average scores on all three models, although the size of the gain varies across model family. This suggests that evidence-localized supervision provides a more targeted adaptation signal than generic span-based self-supervision, while preserving full-context generation. The gains are more visible on several tasks that require locating or integrating evidence across long inputs, such as 2WikiMultihopQA, QASPER, and MultiFieldQA-en. These gains show the benefit of anchoring test-time updates to question-relevant evidence.

\subsection{Efficiency Analysis}

Table~\ref{tab:efficiency_appendix} compares qTTT and EASE-TTT on three profiled LongBench tasks using Qwen3-1.7B. We focus on qTTT because it is the closest adaptation-based baseline: both methods perform query-side test-time adaptation, but use different supervision signals. EASE-TTT improves the average score from 38.0 to 40.1, while increasing the average per-example runtime from 6.7s to 9.1s. This corresponds to a 2.1-point score improvement with an additional 2.4s per example.

The additional cost mainly comes from evidence selection and attention-map supervision. Unlike qTTT, which optimizes a standard next-token prediction loss on sampled spans, EASE-TTT first identifies question-relevant evidence chunks and constructs a soft target over full-context positions. During adaptation, it also extracts and aligns the selected-layer attention distribution with this target, which introduces extra computation beyond the generic span-based objective. Peak GPU memory remains in a comparable range across the profiled runs. Overall, EASE-TTT trades moderate additional latency for better accuracy over qTTT.
\subsection{Ablation Study}

\begin{figure}[t]
    \centering
    \includegraphics[width=\columnwidth]{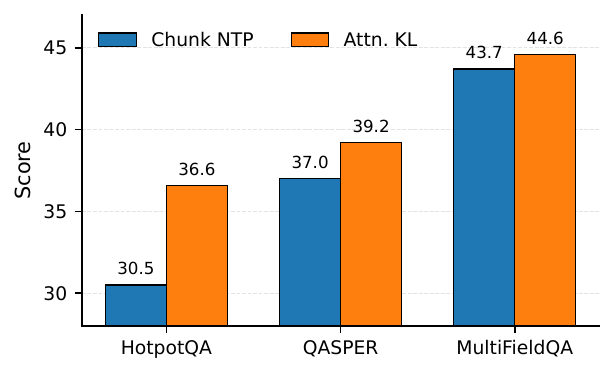}
    \caption{Objective ablation on Qwen3-1.7B. Attn. KL outperforms Chunk NTP, showing the benefit of using selected evidence as explicit attention supervision.}
    \label{fig:objective_ablation}
\end{figure}

\textbf{Loss Objective.}
Figure~\ref{fig:objective_ablation} evaluates the effect of the adaptation objective. Chunk NTP adapts the model on selected evidence chunks using a standard next-token prediction loss, while Attn. KL directly aligns the model's attention distribution with the selected evidence positions. Attn. KL consistently outperforms Chunk NTP on all three tasks, improving HotpotQA from 30.5 to 36.6, QASPER from 37.0 to 39.2, and MultiFieldQA from 43.7 to 44.6. This comparison shows that the benefit of EASE-TTT does not come simply from exposing the model to selected evidence during test-time training. If the selected chunks are used only as ordinary next-token prediction data, the adaptation objective remains weakly connected to the final evidence-access problem. In contrast, Attn. KL converts the selected chunks into an explicit supervision signal over full-context positions. This better matches the goal of EASE-TTT: improving how the model attends to evidence while still generating from the original full context.

\noindent \textbf{Effect of Attention Layer.}
\begin{figure}[t]
    \centering
    \includegraphics[width=\columnwidth]{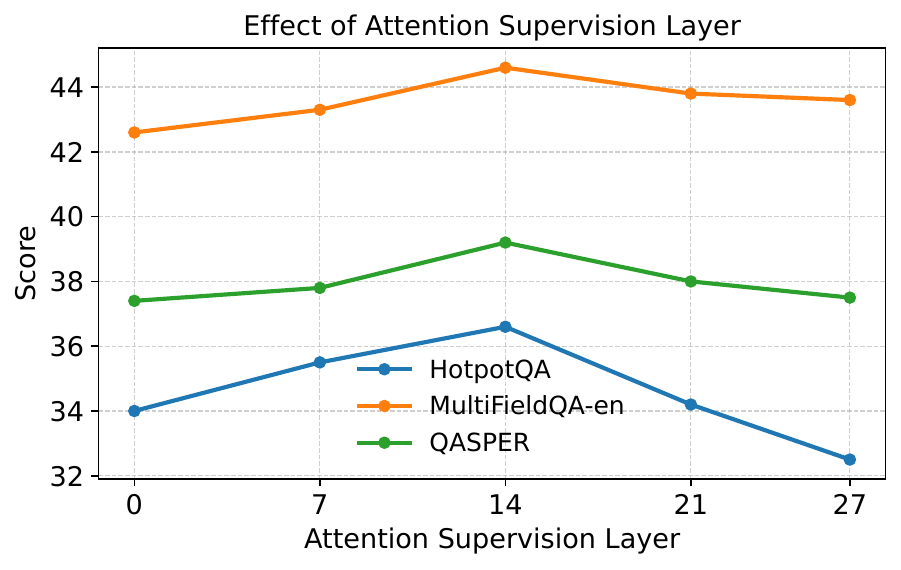}
    \caption{Effect of attention layer on EASE-TTT using Qwen3-1.7B. The results compare different attention layers while keeping all other hyperparameters fixed.}
    \label{fig:layer_sweep}
\end{figure}
Figure~\ref{fig:layer_sweep} studies how the choice of attention supervision layer affects performance. This choice is not merely an implementation detail, because recent layer-wise analyses suggest that different LLM layers play different functional roles. Lower layers are more involved in gathering information from previous tokens, while upper layers increasingly consolidate the gathered information internally~\citep{artzy2024attend}. In addition, intermediate layers can encode stronger task-relevant representations than final layers for downstream tasks~\citep{skean2025layer}.

Our results are consistent with this view. Very early layers are less effective, likely because their attention patterns are still dominated by low-level context gathering rather than question-specific evidence use. The final layer is also not necessarily optimal, since it may be more closely tied to consolidated representations and final prediction. Intermediate layers provide a better trade-off: they are sufficiently contextualized to reflect question-relevant evidence, while still leaving room for the alignment signal to influence subsequent computation. This explains why EASE-TTT benefits more from supervising intermediate attention layers than from supervising the earliest or final layers.

\section{Conclusion}
We studied long-context question answering for smaller language models, where answer-bearing evidence may already be present in the input but not reliably accessed by the model. We proposed EASE-TTT, a within-context retrieval-augmented test-time training framework that localizes evidence chunks and converts them into soft attention supervision for query-side adaptation. Rather than replacing the full context with retrieved chunks, EASE-TTT uses localized evidence to guide lightweight test-time updates while generating the final answer from the original full context. Experiments on six LongBench QA tasks show that EASE-TTT improves over full-context inference, retrieval-only baselines, and qTTT. Ablation results further show that explicit attention alignment is more effective than next-token prediction on selected chunks, suggesting that localized evidence is most useful when it guides how the model attends to the full context rather than only exposing relevant content. These findings highlight evidence-aware test-time adaptation as a promising direction for smaller long-context models.

\section*{Limitations}
This work has several limitations. First, our experiments focus on long-context question answering tasks, where answer-relevant information is usually expected to appear in the input context. Although this setting directly matches our research question, further evaluation is needed to understand how EASE-TTT generalizes to other types of tasks, such as mathematical reasoning, symbolic reasoning, and open-ended generation.

Second, our study mainly evaluates relatively small language models. Since larger models may already have stronger long-context utilization ability, the effect of evidence-guided test-time adaptation may vary across model scales. Future work can examine how the proposed approach behaves on larger models and different model family.

\bibliography{reference}

\appendix

\section{Prompt Templates}

\begin{tcolorbox}[
    enhanced,
    breakable,
    colback=white,
    colframe=blue!70!black,
    colbacktitle=blue!70!black,
    coltitle=white,
    fonttitle=\bfseries,
    title={
        \begin{tabular}{@{}l@{}}
        Instruction: Context-Based Question \\
        Answering
        \end{tabular}
    },
    arc=1.2mm,
    boxrule=0.8pt,
    left=3mm,
    right=3mm,
    top=2.5mm,
    bottom=2.5mm,
    before skip=6pt,
    after skip=6pt
]
Answer the question based on the given context.

\vspace{0.8em}
\noindent\textbf{Context:}

\noindent\texttt{\{context\}}

\vspace{0.8em}
\noindent\textbf{Question:}

\noindent\texttt{\{question\}}

\vspace{0.8em}
\noindent Please provide the final answer only.
\end{tcolorbox}
We use the following prompt template for all context-based question answering experiments. 
The placeholder \texttt{\{context\}} denotes the full input context provided by the benchmark, 
and \texttt{\{question\}} denotes the corresponding question. 
To ensure consistent evaluation, we instruct the model to output only the final answer without 
additional explanations or intermediate reasoning.

\section{Baseline Implementation Details}

\subsection{Full-Context Inference}

We use full-context inference as the base-model baseline. This baseline directly feeds the benchmark-provided context and question into the pretrained model and generates the answer without retrieval, prompt compression, or test-time parameter updates. For a fair comparison, we use the same model checkpoints, tokenizer, prompt template, context truncation, question truncation, maximum answer length, and decoding strategy as EASE-TTT. Specifically, the input context is truncated to at most 32,768 tokens, the question is truncated to at most 1,024 tokens, and the maximum answer length is set to 128 tokens. We use deterministic decoding for all evaluated models. No LoRA adapters are inserted, and all model parameters remain unchanged during inference. 

\subsection{Within-Context RAG}

We implement Within-Context RAG as a retrieval-only baseline that uses the same input context as the original long-context QA instance and does not access any external corpus. For a fair comparison with EASE-TTT, we use the same context preprocessing, tokenizer, truncation limits, prompt template, and decoding settings as our method. The input context is first truncated to at most 32,768 tokens, and the question is truncated to at most 1,024 tokens. The maximum answer length is set to 128 tokens, and deterministic decoding is used.

For retrieval, we segment the truncated context into fixed-length chunks of 512 tokens. We then use BM25 to rank these chunks with the question as the retrieval query and select the top 4 chunks as the retrieved context. The selected chunks are concatenated in their original document order and passed to the base model for answer generation. Within-Context RAG does not access any external documents, does not insert LoRA adapters, and does not perform test-time parameter updates.

\subsection{ICR}

We implement R\&R following the original paper and use the official open-source implementation released by the authors. R\&R combines reprompting and in-context retrieval (ICR) to improve long-context question answering performance. Following the original setup, documents are divided into page-level segments, and the model first performs retrieval by identifying the top-\(k\) most relevant pages for the given question before conducting a second QA step on the abbreviated context. Following the default configuration in the original work, we retrieve the top-\(k=5\) pages during the ICR stage. During reprompting, reminder instruction blocks are periodically inserted throughout the long context to mitigate the lost-in-the-middle effect by reducing the distance between relevant evidence and task instructions. Specifically, reminder prompts are inserted approximately every \(r = 10\text{k}\) tokens following the implementation described in the original paper. The retrieval stage uses the same two-stage retrieval-and-answering pipeline as the original implementation, where the first LLM call retrieves relevant page IDs and the second LLM call performs QA on the abbreviated context constructed from the retrieved pages. Following the original implementation, we use the official prompt templates, retrieval formatting, and hyperparameter settings provided by the authors for all experiments.

\subsection{QTTT}

We implement qTTT following the original paper and use the official open-source implementation released by the authors. Following the original setup, qTTT performs lightweight test-time adaptation on the query projection modules using LoRA adapters rather than updating the full model parameters. During inference, the key and value projections remain frozen, allowing the model to reuse the precomputed KV cache without recomputing full-context representations. Following the default configuration in the original work, qTTT performs \(N_{\text{qTTT}} = 32\) gradient update steps during inference using randomly sampled spans of length \(k = 128\) tokens, with a learning rate of \(1\times10^{-5}\). Test-time optimization is applied only to the query-side attention parameters while all remaining model weights stay frozen. The adaptation objective follows the standard next-token prediction loss computed over sampled context spans, using the optimization procedure and default hyperparameter settings provided in the original implementation. Following the motivation of qTTT, this adaptation strategy is designed to mitigate attention score dilution in long-context reasoning by improving retrieval of relevant context tokens during inference while preserving efficient KV-cache reuse.

\begin{table}[t]
\centering
\small
\caption{Efficiency comparison on three profiled LongBench tasks using Qwen3-1.7B. Time is measured in seconds and memory is measured in GB.}
\label{tab:efficiency_appendix}
\setlength{\tabcolsep}{5pt}
\renewcommand{\arraystretch}{1.05}
\begin{tabular}{llccc}
\toprule
Dataset & Method & Score $\uparrow$ & Time $\downarrow$ & Mem. \\
\midrule
HotpotQA & qTTT & 33.4 & 8.03 & 14.0 \\
         & Ours & 36.6 & 13.80 & 11.6 \\
QASPER   & qTTT & 37.2 & 5.70 & 7.3 \\
         & Ours & 39.2 & 5.73 & 6.5 \\
MultiField & qTTT & 43.3 & 6.35 & 9.0 \\
           & Ours & 44.6 & 7.91 & 7.8 \\
\midrule
Avg. & qTTT & 38.0 & 6.7 & 10.1 \\
     & Ours & 40.1 & 9.1 & 8.6 \\
\bottomrule
\end{tabular}
\end{table}

\subsection{Evidence Selection}

\begin{table}[!t]
\centering
\small
\caption{Effect of evidence source on Qwen3-1.7B. Scores are reported on three LongBench QA tasks.}
\label{tab:evidence_source}
\renewcommand{\arraystretch}{1.05}
\setlength{\tabcolsep}{5pt}
\begin{tabular}{lccc}
\toprule
Source & HotpotQA & QASPER & MultiFieldQA \\
\midrule
BM25 & 36.5 & 38.4 & 44.1 \\
Utility & \textbf{36.6} & \textbf{39.2} & \textbf{44.6} \\
\bottomrule
\end{tabular}
\end{table}

Table~\ref{tab:evidence_source} examines the source of evidence used to construct the attention target. Utility-based selection slightly but consistently improves over BM25 across all three tasks. This suggests that BM25 can retrieve useful lexical matches, while the proposed utility score provides a more task-aligned signal for selecting evidence chunks. Since the utility score measures how much a chunk improves question modeling, it is better aligned with the downstream adaptation objective than purely lexical retrieval. The consistent gains support our use of utility-based evidence selection for evidence-guided test-time training.

\end{document}